\documentclass{article} 
\usepackage{iclr2027_conference,times}
\usepackage{enumitem}
\usepackage[most]{tcolorbox}
\usepackage{amsthm}

\usepackage{hyperref}
\usepackage{url}
\usepackage{booktabs}
\usepackage{graphicx}
\usepackage{multirow}
\usepackage{xcolor}
\usepackage{amsmath,amssymb}
\usepackage{caption}
\usepackage{subcaption}
\usepackage{wrapfig}
\newsavebox{\hpdtablebox}

\title{Interpolated Policy Distillation: A Controllable Continuum Between Off-Policy and On-Policy Distillation}

\iclrfinalcopy
\author{
Youxu Shi\textsuperscript{1 *},
Yifan Sun\textsuperscript{2 *$\dagger$},
Dacheng Yin\textsuperscript{2},
Haomiao Tang\textsuperscript{2},
Guangting Wang\textsuperscript{2},\\
\textbf{Fengyun Rao}\textsuperscript{2},
\textbf{Jing LYU}\textsuperscript{2},
\textbf{Dong Liu}\textsuperscript{1 $\dagger$}\\[0.5em]
\textsuperscript{1}University of Science and Technology of China\\
\textsuperscript{2}WeChat Vision, Tencent Inc.\\
\texttt{syx123@mail.ustc.edu.cn}\\
\texttt{dogeliu@ustc.edu.cn}\\
\texttt{orionsun@tencent.com}
}

\begin{document}

\maketitle
\begingroup
\renewcommand{\thefootnote}{\fnsymbol{footnote}}
\footnotetext[1]{Equal contribution.}
\footnotetext[2]{Corresponding authors.}
\endgroup
\begin{abstract}

Off-policy and on-policy distillation have traditionally been formulated as separate paradigms, each favoring a different property of distillation trajectories. Teacher-generated (off-policy) traces are typically high-quality but lie far from the student's distribution, whereas student-generated (on-policy) rollouts are more learnable but often contain erroneous reasoning. We view these paradigms as the endpoints of a policy continuum and posit that a more effective rollout policy may lie in between. We introduce \textbf{Interpolated Policy Distillation (IPD)}, which defines the next-token distribution at every decoding step as an explicit linear interpolation between the student and teacher distributions. The interpolation operates at the distribution level, token by token, and its coefficient provides direct control over the balance between trajectory quality and student learnability. Naively sampling from this policy would require sequentially querying the teacher at every token and is thus expensive. To make IPD practical, we
accelerate it with a new speculative-decoding rule while exactly preserving the interpolated next-token distribution.At the trajectory level, the resulting rollouts naturally interleave student- and teacher-generated segments. Unlike recent heuristic segment-interleaving methods, however, this interleaving is induced by an exactly realized token-level interpolated policy rather than by hand-designed switching rules.
Across text-only and multimodal reasoning benchmarks, IPD consistently outperforms both endpoint policies (SFT and OPD), their conventional two-stage combination (SFT-then-OPD), and recent heuristic segment-interleaving methods, demonstrating that token-level policy interpolation better balances trajectory quality and student learnability.

\end{abstract}

\section{Introduction}
\begin{figure}[h]
    \centering
    \includegraphics[width=0.9\textwidth]{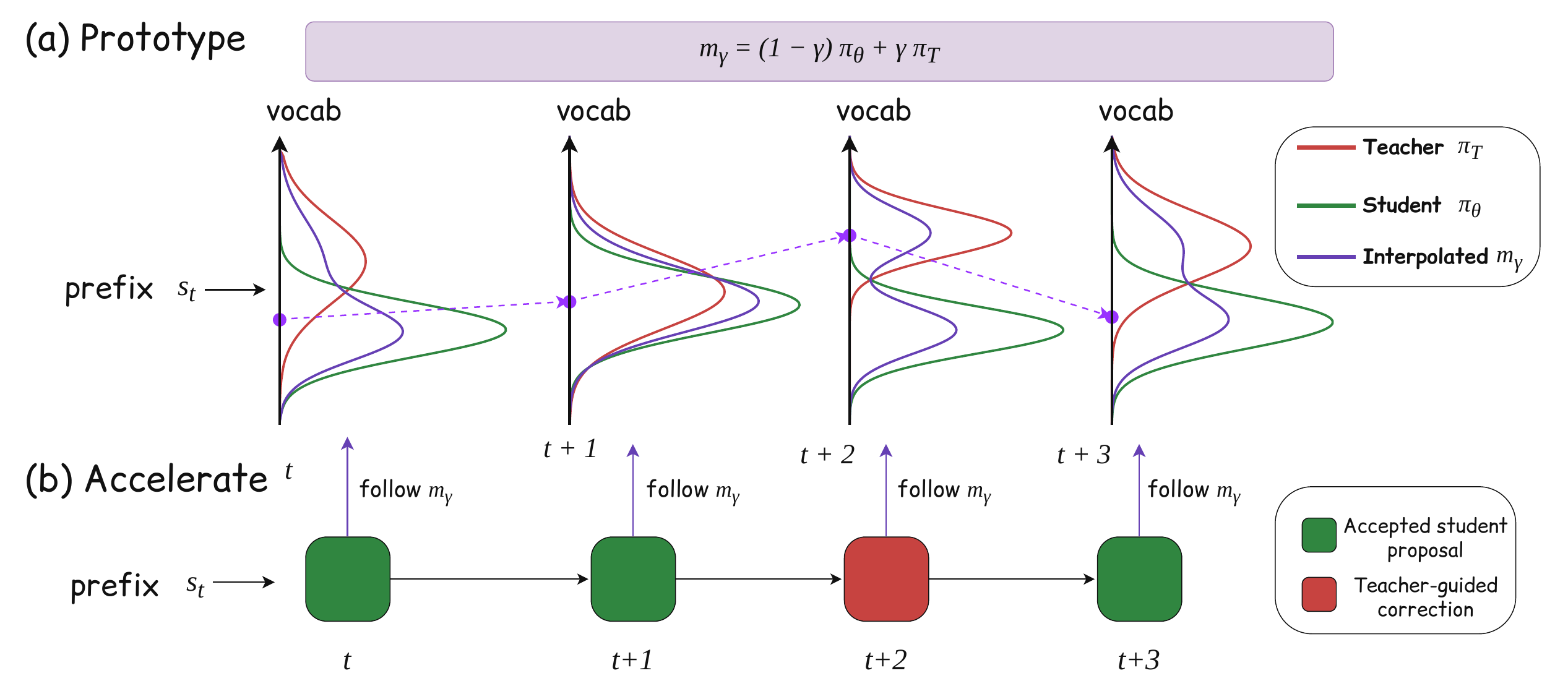}
    \caption{Overview of Interpolated Policy Distillation (IPD). (a) IPD defines a rollout policy \(m_\gamma=(1-\gamma)\pi_\theta+\gamma\pi_T\), interpolating between the student and teacher distributions at each prefix to balance trajectory quality and student learnability. (b) Trajectory generation is accelerated by a tailored speculative sampling scheme: at each step the student proposal is accepted when possible (green), and otherwise replaced by a teacher-guided correction (red). In both cases the emitted token follows exactly \(m_\gamma(\cdot \mid s_t)\), so the speedup is lossless with respect to the rollout distribution.
}
    \label{fig:hpd_mechanism}
\end{figure}

Trajectories matter in distilling large language models (LLMs): what a student learns is determined by the token sequences on which it is trained, and distillation methods differ fundamentally in \emph{who generates} them. Off-policy distillation~\citep{Hinton2015, kim2016sequence, shridhar2023distilling} trains on teacher-generated or fixed-corpus trajectories, whereas on-policy distillation (OPD)~\citep{agarwal2024onpolicy, lu2025onpolicydistillation} trains on the student's own rollouts. This choice creates a tension between two properties of a trajectory: its \emph{quality}, whether it constitutes a correct and valid reasoning process, and its \emph{learnability}, how readily the current student can absorb its supervision. Teacher-generated trajectories tend to be high-quality but induce a train--inference mismatch~\citep{lin2020autoregressive}: the student is supervised on prefixes that it would rarely visit on its own. Student-generated rollouts remove this mismatch, but for weak students they drift into erroneous or degenerate prefixes, and a large teacher--student gap can further destabilize training through extreme importance signals, with high local agreement failing to translate into successful reasoning~\citep{xin2026escapingklagreementtrap}. Neither endpoint, therefore, is guaranteed to strike the best balance between quality and learnability.

This tension suggests that the choice between off-policy and on-policy distillation need not be binary. We instead view the student and teacher policies as the endpoints of a \emph{continuum} of rollout policies. Moving toward the teacher can improve trajectory quality by correcting unsuitable student decisions, whereas moving too far increases the distributional gap that the student must bridge. The most effective trajectories may therefore come from an \emph{intermediate} policy---one that incorporates enough teacher guidance to preserve quality while remaining close enough to the student's evolving distribution to be learnable. This raises our central question: \textbf{can we construct a rollout policy at a controllable, precisely characterized position between the student and teacher, thereby directly balancing trajectory quality and student learnability?}

We answer this with \textbf{Interpolated Policy Distillation (IPD)}, which specifies the rollout policy at the distribution level: at each decoding step, the next-token distribution $m_\gamma$ is a linear interpolation of the student and teacher distributions, governed by a single coefficient $\gamma\in[0,1]$. Larger $\gamma$ moves the policy toward the teacher, whereas smaller $\gamma$ keeps it near the student; at $\gamma=0$ IPD reduces to vanilla OPD, and at $\gamma=1$ it rolls out from the teacher as in off-policy distillation. We formalize $m_\gamma$ as a policy continuum in Section~\ref{sec:continuum}, and show how to sample from it exactly and efficiently in Section~\ref{sec:method}.

\begin{figure}[h]
    \centering
    \includegraphics[width=0.8\textwidth]{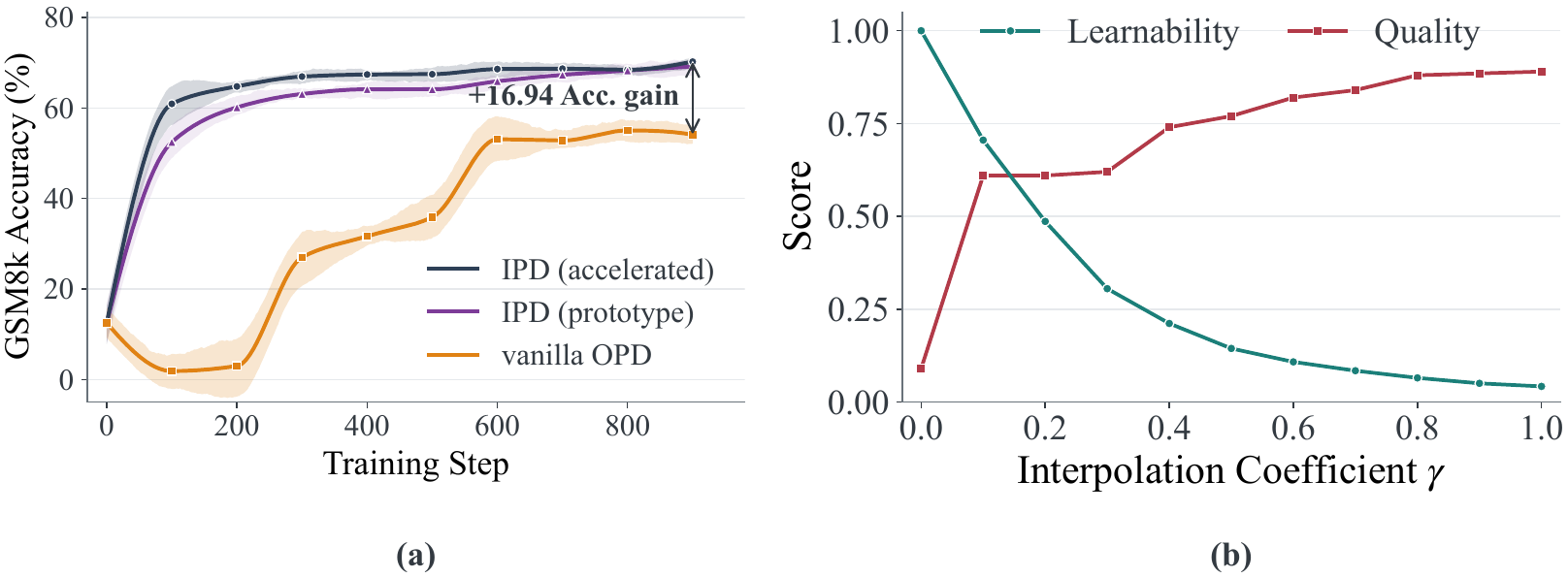}
    \caption{
    (a) On a representative setting (Qwen3-4B $\rightarrow$ Qwen3-0.6B-Base, evaluated on GSM8K), both the IPD prototype and its accelerated version outperform OPD by clear margins. In particular, the prototype applies the vanilla OPD objective to every token ,yet it recovers most of the gain, indicating the gains mainly stem from the interpolated policy. (b) Effect of the interpolation coefficient $\gamma$: a proxy for learnability falls monotonically as $\gamma$ grows, while rollout quality (scored by an LLM judge) rises, exposing the trade-off that $\gamma$ balances.}
    \label{fig:gsm8k_curves_bars_row}
\end{figure}

In one representative setting, a direct-sampling prototype of IPD substantially outperforms vanilla OPD (Fig.~\ref{fig:gsm8k_curves_bars_row}), providing controlled evidence that the interpolated rollout policy itself can improve distillation. Direct sampling from $m_\gamma$, however, requires a sequential teacher query at every token and is therefore expensive. To make IPD practical, we develop a speculative-sampling rule in which the student proposes tokens in blocks and a single teacher pass per block either accepts them or replaces them with a teacher-guided correction, reducing sequential teacher decoding while preserving the interpolated next-token distribution exactly. We state this distributional guarantee in Section~\ref{sec:accelerated} and prove it in Appendix~\ref{app:exact_sampling}.

At the trajectory level, accelerated IPD manifests as an interleaving of accepted student proposals and teacher-guided corrections. Recent teacher-intervention methods (discussed in Section~\ref{sec:related}), produce similar-looking trajectories but decide when and how much the teacher intervenes through hand-designed switching rules; IPD instead specifies the target token-level distribution $m_\gamma$ first and derives a distribution-preserving sampling procedure from it. Notably, the prototype performs no explicit interleaving at all, yet exhibits the same improvement trend as accelerated IPD in Fig.~\ref{fig:gsm8k_curves_bars_row}, indicating that the gains stem from the interpolated policy rather than from the interleaving pattern.

Because the sampler records whether each emitted token was an accepted proposal or a teacher-guided correction, a single rollout can combine both forms of supervision: IPD applies direct likelihood supervision to teacher-guided corrections and OPD supervision to accepted student proposals, rather than separating the two into successive training stages. Across text-only and multimodal reasoning benchmarks, and for both base and instruction-tuned students, IPD consistently outperforms not only both endpoint baselines---off-policy SFT and vanilla OPD---but also their conventional two-stage combination, SFT-then-OPD, and recent segment-interleaving methods. These results show that IPD's gains arise not merely from combining off-policy and on-policy supervision, but from combining them through an explicitly defined token-level policy interpolation.

\paragraph{Contributions.}
\textbf{(1) A unified policy continuum.} We formulate off-policy distillation and OPD as the endpoints of a single policy continuum, exposing intermediate rollout policies as an explicit design space for balancing trajectory quality and student learnability.
\textbf{(2) Interpolated Policy Distillation.} We instantiate this continuum as a per-token interpolation of the student and teacher distributions, and accelerate it with a speculative-sampling rule that preserves the interpolated distribution exactly. The same accept--reject decisions also determine which supervision applies at each token.
\textbf{(3) Empirical validation.} Across text-only and multimodal reasoning benchmarks and both base and instruction-tuned students, IPD consistently outperforms off-policy SFT, vanilla OPD, their conventional two-stage combination, and recent segment-interleaving methods.

\section{A Trajectory-Policy Continuum}
\label{sec:continuum}

This section formalizes the trajectory-policy continuum and reviews related distillation methods through this unified perspective. Throughout, $\pi_\theta$ and $\pi_T$ denote the student and teacher policies and $s_t=(x,y_{<t})$ denotes the prefix state at decoding step $t$.

\subsection{The Policy Continuum}
\label{sec:unified_continuum}

At each decoding step, given prefix $s_t=(x,y_{<t})$, we define the next-token distribution
\begin{equation}
m_\gamma(\cdot\mid s_t)
=
(1-\gamma)\,\pi_\theta(\cdot\mid s_t)
+
\gamma\,\pi_T(\cdot\mid s_t),
\qquad \gamma\in[0,1],
\label{eq:interpolated_policy}
\end{equation}
which induces the trajectory distribution $P_\gamma(y\mid x)=\prod_{t=1}^{|y|}m_\gamma(y_t\mid s_t)$. Note that $P_\gamma\neq(1-\gamma)P_{\pi_\theta}+\gamma P_{\pi_T}$: interpolating the conditional distributions at every prefix is not the same as interpolating the two complete trajectory distributions, which would amount to choosing one of the two policies once and generating the entire trajectory from it. The coefficient $\gamma$ sets the rollout policy's position on the continuum: $\gamma=0$ and $\gamma=1$ recover the student and teacher rollout policies, respectively, while $\gamma\in(0,1)$ defines the intermediate policies explored by IPD.

\subsection{The Two Endpoints and the Interior}
\label{sec:endpoints}

{Motivated by trust region theory~\citep{pmlr-v37-schulman15}}, we assess the learnability of a rollout policy through its proximity to the student's own distribution. We measure this proximity by the total variation distance $D_{\mathrm{TV}}(u,w)=\frac{1}{2}\sum_v|u(v)-w(v)|$ between two distributions over the vocabulary. For $m_\gamma$ it admits an exact form at every prefix:
\begin{equation}
\begin{aligned}
D_{\mathrm{TV}}\!\left(m_\gamma,\,\pi_\theta\right)
=
\gamma\,D_{\mathrm{TV}}\!\left(\pi_\theta,\,\pi_T\right), \
D_{\mathrm{TV}}\!\left(m_\gamma,\,\pi_T\right)
=
(1-\gamma)\,D_{\mathrm{TV}}\!\left(\pi_\theta,\,\pi_T\right).
\end{aligned}
\label{eq:policy_continuum_tv}
\end{equation}
The two distances sum to $D_{\mathrm{TV}}(\pi_\theta,\pi_T)$ and are linear in $\gamma$, so the coefficient places the rollout policy at a precisely known position between the two endpoints rather than merely somewhere between them.
Eq.~\eqref{eq:policy_continuum_tv} is a statement about individual decoding steps, and it should not be read as bounding the distance between the resulting trajectory distributions. Because every token is conditioned on the preceding ones, a single deviation alters every subsequent conditional. A small $\gamma$ therefore does not imply that trajectories sampled from $m_\gamma$ stay close to those the student would have produced: modest teacher influence per token can accumulate into substantially different reasoning paths, which is what makes the interior of the continuum worth exploring even for small $\gamma$.

\paragraph{The teacher endpoint.}
At $\gamma=1$ the rollout policy is $\pi_T$, attaining the largest distance from the student. Teacher rollouts typically provide high-quality reasoning~\citep{Hinton2015,kim2016sequence,shridhar2023distilling}, but they visit prefixes and select continuations that are unlikely under the student. High trajectory quality therefore does not imply high learnability: the supervision the teacher provides may be difficult for the student to absorb when the teacher--student gap is large.

\paragraph{The student endpoint.}
At $\gamma=0$ the rollout policy is $\pi_\theta$ itself, so the trajectories match the states the student encounters at inference. This eliminates rollout-distribution shift but offers no guarantee of quality. A weak student may assign substantial probability to erroneous reasoning steps and enter invalid or degenerate prefixes~\citep{li2026rethinking}, and because each token is conditioned on the preceding ones, early mistakes compound and steer the rollout increasingly far from a valid reasoning path. A trajectory can therefore be fully on-policy yet provide a poor basis for distillation, with local agreement coexisting with globally unsuccessful reasoning, as observed in the agreement trap~\citep{xin2026escapingklagreementtrap}.

\paragraph{Why the interior?}
The teacher endpoint thus sacrifices learnability, whereas the student endpoint sacrifices quality. Intermediate values $\gamma\in(0,1)$ introduce teacher guidance without moving all the way to the teacher endpoint. As shown in Figure~\ref{fig:gsm8k_curves_bars_row}(b), trajectory quality, evaluated by an LLM judge, and learnability, measured by proximity to $\pi_\theta$ (metric is detailed in Appendix~\ref{app:metric_quality_learnability}), move in opposite directions as $\gamma$ varies, empirically revealing a trade-off between the two properties. This motivates a policy that balances them better than either endpoint. How to sample from $m_\gamma$ efficiently while preserving its specified distribution is the subject of Section~\ref{sec:method}.

\subsection{Related Work}
\label{sec:related}


\paragraph{Hybrid student--teacher rollouts.}
Several recent methods combine student and teacher generation within the same trajectory. SKD~\citep{xu2025speculative} replaces student proposals outside the teacher's top-$K$ predictions; Relay-OPD~\citep{xu2026pass} detects failed prefixes and temporarily transfers generation to the teacher; CA-OPD~\citep{li2026caopdconfidenceawareonpolicydistillation} corrects unreliable tokens according to teacher confidence and progressively relaxes intervention; and, in the related RL setting, MInTRL~\citep{chen2026mintrloffpolicyinterventionboost} inserts short judge-identified corrections before returning control to the student. These methods combine the two models through method-specific rules that determine when and how much the teacher intervenes, leaving the induced next-token distribution implicit. IPD also produces hybrid student--teacher rollouts, but defines the combination at the distribution level: it first specifies $m_\gamma$ in Eq.~\eqref{eq:interpolated_policy} and then derives a sampling procedure that preserves this distribution. The resulting interleaving is therefore governed by an explicit token-level policy rather than by hand-designed intervention rules.


\section{Interpolated Policy Distillation}
\label{sec:method}

\subsection{The IPD Prototype and Its Limitations}
\label{sec:prototype}

\paragraph{Trajectory rollout policy.}
Following the discussion in Section~\ref{sec:continuum}, an effective rollout policy should
balance trajectory quality and student learnability. We therefore adopt the interpolated policy
$m_\gamma$ of Eq.~\ref{eq:interpolated_policy} as the rollout policy, and sample directly from
it by querying both models at every decoding step. 
Without loss of generality, we choose $\gamma=0.1$ unless otherwise noted (see Section~\ref{sec:ablations} for an ablation over $\gamma$)

\paragraph{Supervision.}
We supervise the prototype with the vanilla OPD objective, unchanged: a
PPO-style surrogate whose per-token reverse-KL term is estimated by the
$k_1$ estimator~\citep{schulman2020kl}.
This is reasonable at the small $\gamma$ we use: by
Eq.~\eqref{eq:policy_continuum_tv}, $m_\gamma(\cdot\mid s_t)$ stays within
$\gamma D_{\mathrm{TV}}(\pi_\theta(\cdot\mid s_t),\pi_T(\cdot\mid s_t))$ of
the student at every prefix, close enough for student-side supervision to
apply.
A mismatch nonetheless remains: the importance ratio is formed against the
student snapshot $\pi_{\theta_{\mathrm{old}}}$ that collected the rollout,
exactly as in on-policy training, so it corrects for the staleness of the
snapshot but not for the tokens having been drawn from $m_\gamma$ rather
than from $\pi_{\theta_{\mathrm{old}}}$ itself.
This discrepancy is $O(\gamma)$ in total variation, and we leave it
uncorrected.

\paragraph{The prototype already improves over vanilla OPD.}
We train Qwen3-0.6B-Base with the prototype, using the same training
hyperparameters as vanilla OPD (Section~\ref{sec:experiments}).
As shown in Figure~\ref{fig:gsm8k_curves_bars_row}(a), it substantially
outperforms vanilla OPD, converging faster and reaching higher final
accuracy.
Since the prototype inherits the vanilla OPD objective unchanged, the two
runs share their objective, data, and number of optimization steps, and
differ only in the policy that generated the rollouts.
The gain is therefore attributable to the interpolated policy alone: moving
into the interior of the continuum helps even before any change to the
supervision.


\paragraph{Two limitations.}
Training the prototype nonetheless reveals two limitations, both stemming from its sampling directly
from $m_\gamma$.

\emph{- Inefficiency.} Sampling from $m_\gamma$ requires a teacher forward pass at every decoding
step, so teacher computation is strictly sequential and cannot be amortized across positions;
maintaining two generation engines simultaneously adds further coordination overhead.

\emph{- Undifferentiated supervision.} 
\emph{Undifferentiated supervision.}
Rollouts from $m_\gamma$ contain both tokens the student would plausibly generate itself and tokens the student would rarely generate on its own (but which the teacher favors).
While the former suit an on-policy divergence objective, the latter do not: on these tokens the log ratios in the divergence objective have large magnitude and can destabilize optimization, so a likelihood-based objective is preferable. Direct sampling from $m_\gamma$ does not indicate which kind each token is,
so the prototype applies a single objective throughout.



Sections~\ref{sec:accelerated} and~\ref{sec:supervision} address these limitations in turn.
Speculative sampling removes the per-token sequential teacher query and, at no additional cost,
labels each emitted token as an accepted student proposal or a teacher-guided correction; these
labels in turn make differentiated supervision possible.

\subsection{Speculative Sampling from the Interpolated Policy}
\label{sec:accelerated}

Speculative sampling~\citep{pmlr-v202-leviathan23a} is a general recipe for LLM inference acceleration: given a cheap proposal distribution and any target distribution, one drafts tokens from the proposal and then accepts or corrects them, obtaining exact samples from the target while querying it only once per block of drafted tokens. 
We take the student as the proposal and the interpolated
policy $m_\gamma$ of Eq.~\ref{eq:interpolated_policy} as the target policy, i.e., 
the verifier.

\begin{figure}[t]
    \centering
    \includegraphics[width=0.95\textwidth]{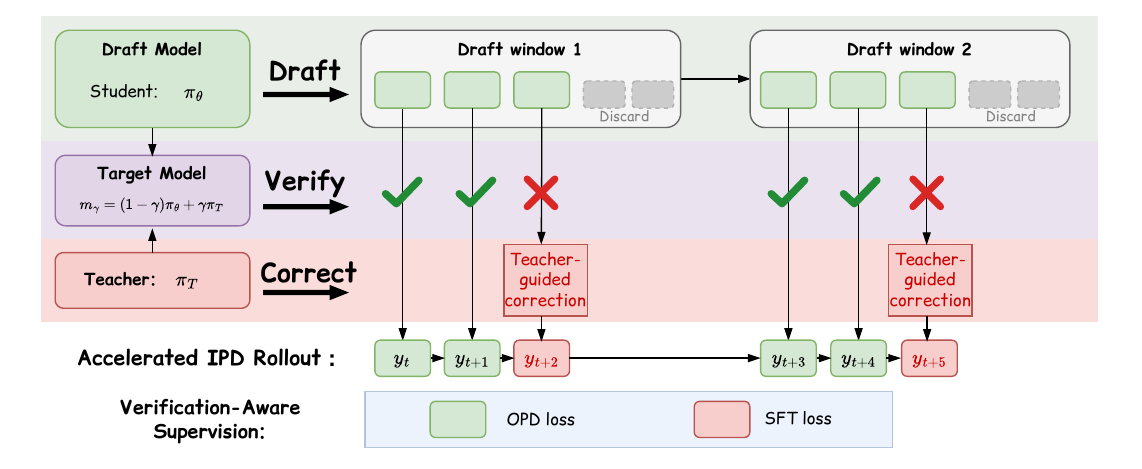}
    \caption{Accelerated IPD: speculative sampling (Section~\ref{sec:accelerated}) with verification-aware supervision (Section~\ref{sec:supervision}). The student drafts $k$ tokens sequentially and the teacher scores the block in a single forward pass, giving the target policy $m_\gamma$ at every draft position. Proposals are accepted with probability $a_\gamma$ (Eq.~\ref{eq:accept_pro}); the first rejected one is replaced by a teacher-guided correction drawn from the residual $r_\gamma$ (Eq.~\ref{eq:ipd_residual}) and the draft suffix is discarded. Every generated token therefore follows $m_\gamma(\cdot\mid s_t)$ exactly.The same decisions determine supervision: accepted proposals (green) receive the OPD objective (Eq.~\ref{eq:ipd_accepted_loss}), corrections (red) direct likelihood supervision (Eq.~\ref{eq:ipd_correction_loss}).}
    \label{fig:hpd_overview}
\end{figure}

\paragraph{Trajectory construction.}
Starting from the current prefix, the student drafts $k$ tokens autoregressively, and the verifier
evaluates all draft positions in a single forward pass (Figure~\ref{fig:hpd_overview}). Writing
$p(v)=\pi_\theta(v\mid s)$ and
$q(v)=\pi_T(v\mid s)$ for the student and teacher distributions at a draft prefix $s$, a proposed
token $v$ is accepted with probability
\begin{equation}
a_\gamma(v\mid s)
=
\min\left\{1,\frac{m_\gamma(v\mid s)}{p(v)}\right\}
=
\min\left\{1,\,1-\gamma+\gamma\frac{q(v)}{p(v)}\right\}.
\label{eq:accept_pro}
\end{equation}
The ratio is evaluated only for proposed tokens, for which $p(v)>0$, and all acceptance
probabilities are computed in parallel with independent acceptance draws.

IPD retains the longest accepted prefix of the draft. At the first rejection, the rejected
proposal is replaced by a token drawn from the residual distribution at the same prefix,
\begin{equation}
r_\gamma(v\mid s)
=
\frac{[m_\gamma(v\mid s)-p(v)]_+}{\sum_u[m_\gamma(u\mid s)-p(u)]_+},
\qquad [a]_+=\max(a,0).
\label{eq:ipd_residual}
\end{equation}

The remaining draft tokens are discarded, since they were conditioned on the rejected proposal. After a rejection or a full acceptance, the draft window advances to the next unfilled position, and drafting resumes from the updated prefix until the trajectory is complete.
We record the outcome of each step as $z_t\in\{0,1\}$, with $z_t=1$ if $y_t$ is an accepted student proposal and $z_t=0$ if it is a teacher-guided correction.

\paragraph{What $\gamma$ controls.}
Since any two token distributions both sum to one, the total variation
distance can equivalently be written in positive-part form,
$D_{\mathrm{TV}}(u,w)=\sum_v[u(v)-w(v)]_+$, which isolates exactly the tokens
on which one distribution places more mass than the other. In these terms, a
proposal is rejected only when the student assigns it higher probability than
the teacher, and
\begin{equation}
\Pr(z_t=0\mid s_t=s)
=
D_{\mathrm{TV}}\!\left(p,m_\gamma(\cdot\mid s)\right)
=
\gamma\,D_{\mathrm{TV}}(p,q).
\label{eq:ipd_intervention}
\end{equation}
Moreover, since $m_\gamma-p=\gamma(q-p)$, the coefficient $\gamma$ cancels in
Eq.~\eqref{eq:ipd_residual}, giving
\begin{equation}
r_\gamma(v\mid s)=\frac{[q(v)-p(v)]_+}{D_{\mathrm{TV}}(p,q)}=r_1
\label{eq:ipd_residual_independence}
\end{equation}
whenever rejection has positive probability. Thus $\gamma$ sets \emph{how
often} the teacher intervenes, while \emph{what} the intervention does is
determined by where the teacher and student disagree, independently of
$\gamma$: corrections concentrate at prefixes of large
$D_{\mathrm{TV}}(p,q)$ and are supported only on tokens the teacher favors
more than the student.
Equation~\eqref{eq:ipd_intervention} also explains the efficiency gain, as the
expected number of consecutive accepted tokens scales as
$1/\bigl(\gamma D_{\mathrm{TV}}(p,q)\bigr)$: small $\gamma$ yields long
accepted runs and correspondingly few teacher forward passes.
More details are shown in Appendix~\ref{app:exact_sampling}.

\subsection{Verification-Aware Supervision}
\label{sec:supervision}

Recall that the prototype had to apply a single objective to every token of a
rollout (Section~\ref{sec:prototype}), not because one objective suits them
all, but because sampling from $m_\gamma$ returns only the token drawn and
not which kind it is.
Speculative sampling removes this obstacle at no additional cost: the label
$z_t$ records whether $y_t$ was an accepted student proposal or a
teacher-guided correction.


\paragraph{Accepted proposals.}
Accepted tokens tend to lie in the student's own high-probability region, preserving a strong on-policy character. Acceptance also naturally
attenuates extreme log-ratio signals, making these tokens well suited
to OPD supervision. Let $\pi_{\theta_{\mathrm{old}}}$ denote the student policy used to collect the
rollout. We use the per-token reverse-KL signal given by the $k_1$
estimator~\citep{schulman2020kl} as a fixed advantage,
\begin{equation}
\widehat{A}^{\mathrm{KD}}_t
=
\log\pi_T(y_t\mid s_t)-\log\pi_{\theta_{\mathrm{old}}}(y_t\mid s_t),
\label{eq:ipd_kd_advantage}
\end{equation}
and optimize a clipped surrogate
\begin{equation}
\ell^{\mathrm{acc}}_t(\theta)
=
-\min\left\{
\rho_t(\theta)\widehat{A}^{\mathrm{KD}}_t,\;
\operatorname{clip}\!\left(\rho_t(\theta),1-\epsilon,1+\epsilon\right)\widehat{A}^{\mathrm{KD}}_t
\right\},
\label{eq:ipd_accepted_loss}
\end{equation}
where $\rho_t(\theta)=\pi_\theta(y_t\mid s_t)/\pi_{\theta_{\mathrm{old}}}(y_t\mid s_t)$ and
$\epsilon$ is the clipping coefficient.

\paragraph{Residual corrections.}
Corrections are drawn from $[q-p]_+$ by Eq.~\ref{eq:ipd_residual_independence} and are thus, by
construction, tokens the student under-weights relative to the teacher. On these tokens a ratio-based objective can be unstable, since
$\widehat{A}^{\mathrm{KD}}_t$ grows sharply as the student probability decreases. We instead apply direct likelihood supervision,
\begin{equation}
\ell^{\mathrm{rej}}_t(\theta)=-\log\pi_\theta(y_t\mid s_t),
\label{eq:ipd_correction_loss}
\end{equation}
whose gradient magnitude remains bounded and which supplies a mass-covering signal exactly where
the student fails to cover the teacher---the regime in which a mode-seeking divergence provides
little pressure.

\paragraph{Objective.}
Let $\mathcal{T}$ denote the set of generated positions in a rollout. The overall objective is
\begin{equation}
\mathcal{L}_{\mathrm{IPD}}(\theta)
=
\frac{1}{|\mathcal{T}|}
\sum_{t\in\mathcal{T}}
\left[
z_t\,\ell^{\mathrm{acc}}_t(\theta)+(1-z_t)\,\ell^{\mathrm{rej}}_t(\theta)
\right].
\label{eq:ipd_objective}
\end{equation}
SFT-style and OPD-style supervision are thereby combined within a single rollout, with the
speculative decisions determining which signal applies at each token. At $\gamma=0$ every
proposal is accepted and Eq.~\ref{eq:ipd_objective} reduces exactly to vanilla OPD; as $\gamma$
increases, corrections become more frequent and an increasing fraction of supervision takes
the form of likelihood training on teacher-corrected tokens.

Because speculative sampling preserves $m_\gamma$, the accelerated
variant trains on the same rollout distribution as the prototype of
Section~\ref{sec:prototype} and differs from it only in supervision. As shown in Figure~\ref{fig:gsm8k_curves_bars_row}(a), it matches and slightly exceeds the prototype,
indicating that the additional gain comes from differentiated supervision rather than from any
change in the trajectories themselves.
\begin{table*}[t]
\centering
\caption{Comparison of different distillation methods across text reasoning benchmarks. For each benchmark, we report the accuracy (Avg@32).}
\label{tab:main_results}
\resizebox{\textwidth}{!}{
\setlength{\tabcolsep}{3pt}
\begin{tabular}{l cccccc ccccc}
\toprule
& \multicolumn{6}{c}{Qwen3-1.7B-Base $\leftarrow$ Qwen3-4B} & \multicolumn{5}{c}{Qwen3-0.6B-Base $\leftarrow$ Qwen3-4B} \\
\cmidrule(lr){2-7} \cmidrule(lr){8-12}
\textbf{Method} & AIME24 & AIME25 & AIME26 & AMC23 & Olympiad & Avg. & GSM8k & MATH-500 & AMC23 & Olympiad & Avg.\\
\midrule
Student & 5.42 & 2.92 & 1.25 & 23.44 & 17.89 & 10.18 & 12.53 & 7.43 & 7.19 & 7.91 & 8.77 \\
SFT & 4.92 & 2.04 & 1.89 & 25.01 & 19.27 & 10.63 & 34.15& 16.65 & 5.31 & 4.69 & 15.20 \\
GRPO & 5.42 & 1.67 & 3.75 & 33.44 & 20.67 & 12.99 & 60.54 & 37.93 & 20.25 & 8.76 & 31.87  \\
vanilla OPD & 5.83 & 6.67 & 5.00 & 33.75 & 22.31 & 14.71 & 53.08 & 31.52 & 16.56 & 10.17 & 27.83 \\
SKD & 4.28 & 5.96 & 3.76 & 26.71 & 19.67 & 12.08 & 51.46 & 31.29 & 18.27 & 12.19 & 28.30 \\
Relay-OPD & 10.28 & \textbf{7.12} & 5.89 & 32.17 & 22.89 & 15.67 & 67.46 & 43.15 & 22.81 & 14.41 & 36.96 \\
\textbf{IPD (ours)} & \textbf{10.58} & 6.88 & \textbf{6.25} & \textbf{35.31} & \textbf{24.17} & \textbf{16.64} & \textbf{70.02} & \textbf{44.68} & \textbf{24.12} & \textbf{15.06} & \textbf{38.47}\\
\bottomrule
\end{tabular}
}
\end{table*}

\section{Experiments}
\label{sec:experiments}



\subsection{Experimental Setup}
\label{sec:experimental-setup}

\paragraph{Text-only reasoning.}
For weak student initializations, we use Qwen3-0.6B-Base and Qwen3-1.7B-Base as students, with Qwen3-4B as the teacher~\citep{yang2025qwen3}. For a stronger initialization, we use DeepSeek-R1-Distill-Qwen-1.5B~\citep{deepseekai2025deepseekr1incentivizingreasoningcapability} as the student and Skywork-OR1-7B~\citep{he2025skywork} as the teacher.
All three settings use the same 25,600 math questions sampled from OpenThoughts3~\citep{guha2025openthoughtsdatarecipesreasoning}.
Depending on the setting, we evaluate on GSM8K~\citep{cobbe2021gsm8k}, MATH-500~\citep{lightman2023lets}, AMC 2023~\citep{amc2023}, OlympiadBench~\citep{he2024olympiadbench}, and AIME 2024--2026~\citep{jia2024aime,aime2025}.
\paragraph{Multimodal reasoning.}
We consider two settings using Qwen3-VL-8B-Instruct as the teacher. The first uses Qwen3-VL-2B-Instruct~\citep{Qwen3-VL} as the student. The second removes the student's last three decoder layers to create a more challenging initialization, allowing us to assess capability recovery after pruning.
Both settings are trained on Innovator-VL-RL-172K~\citep{wen2026innovator} and evaluated on MathVision~\citep{wang2024measuring}, MathVista~\citep{lu2024mathvista}, MMStar~\citep{chen2024we}, WeMath~\citep{qiao2025we}, and MMMU-Pro~\citep{yue2025mmmu}.
\begin{figure*}[t]
\centering
\begin{minipage}[t]{0.35\textwidth}
    \centering
    \includegraphics[width=\linewidth]{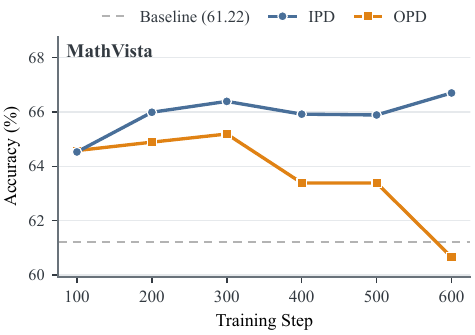}
    \caption{Performance on MathVista.}
    \label{fig:first}
\end{minipage}
\hspace{0.06\textwidth}%
\begin{minipage}[t]{0.38\textwidth}
    \centering
    \includegraphics[width=\linewidth]{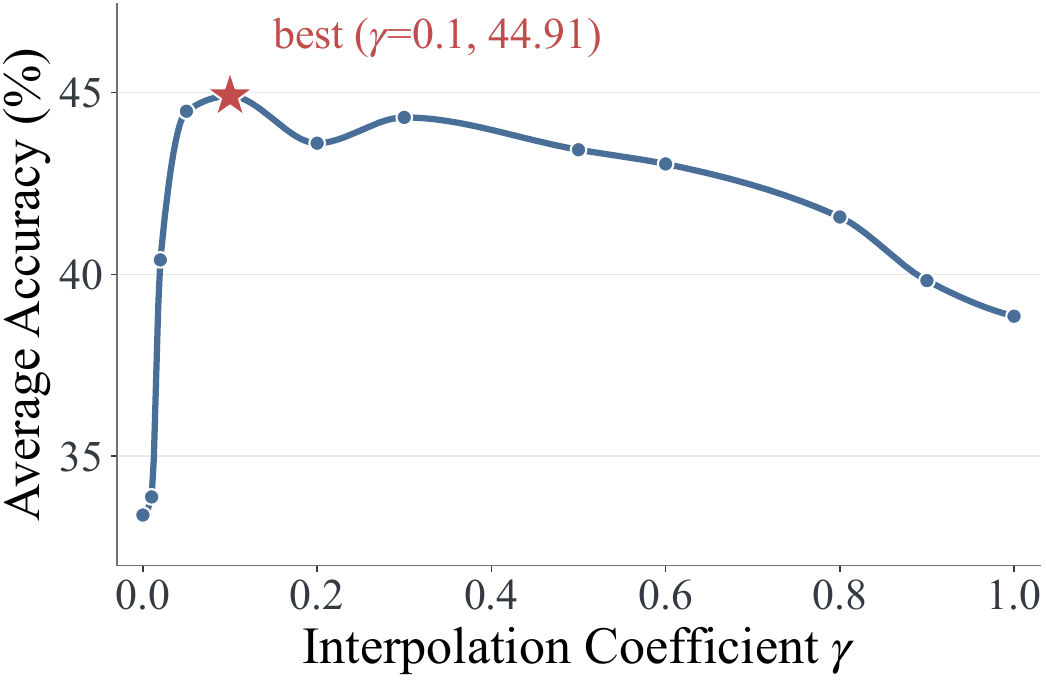}
    \caption{Ablation results averaged on GSM8k, MATH-500 and AMC.}
    \label{fig:second}
\end{minipage}
\end{figure*}
\subsection{Main Results}
\subsubsection{Unimodal Distillation}
Table~\ref{tab:main_results} reports text reasoning results with
two base students distilled from Qwen3-4B.
IPD achieves the highest average accuracy in both settings,
improving over vanilla OPD by 1.93 points for Qwen3-1.7B-Base
and 10.64 points for Qwen3-0.6B-Base.
Among methods that incorporate teacher intervention,
SKD shows mixed results relative to vanilla OPD,
whereas Relay-OPD consistently improves average accuracy.
IPD further surpasses Relay-OPD by 0.97 and 1.51 points,
respectively, and achieves the best performance on eight of
the nine benchmark--student combinations. In fact, results from Relay-OPD are obtained
using the recommended default configurations from its
respective works.
As shown in the Appendix~\ref{app:relay_suffers}, Relay-OPD suffers from training
collapse when run under exactly the same setting as ours, highlighting its sensitivity
to the training setup.
These results support the effectiveness of explicitly controlling
the rollout policy through interpolation.

We additionally evaluate DeepSeek-R1-Distill-Qwen-1.5B
to examine whether these gains extend to a strong student
that has undergone extensive post-training.
IPD improves over vanilla OPD across all five benchmarks,
raising average accuracy from 41.37 to 42.18,
with detailed results provided in the Appendix~\ref{app:deepseek}.
Together, these results demonstrate the effectiveness of IPD
across both base and extensively post-trained students.

\subsubsection{Multimodal Distillation}

Table~\ref{tab:multimodal_results} shows that IPD delivers substantial 
gains not only on LLM benchmarks but also in multimodal settings. 
Across multiple multimodal reasoning benchmarks, IPD consistently 
outperforms vanilla OPD.
Notably, under prolonged training, vanilla OPD exhibits 
progressive performance degradation, whereas IPD maintains stable 
performance and converges to a stronger point, as shown in 
Figure~\ref{fig:first}.
To evaluate IPD under a more challenging initialization,
we remove the final three decoder layers from
Qwen3-VL-2B-Instruct and use distillation to recover its capabilities.
The advantage of IPD becomes considerably larger in this setting:
it achieves an average accuracy of 44.64 compared with 27.90
for OPD, nearly restoring the original unpruned model's
performance (45.87).
This demonstrates that IPD can effectively recover capabilities
lost through pruning without an additional recovery stage.

\begin{table*}[t]
\centering
\caption{Results across multimodal reasoning benchmarks. Avg@32 accuracy is reported.}
\label{tab:multimodal_results}
\resizebox{0.75\textwidth}{!}{%
\setlength{\tabcolsep}{5pt}
\begin{tabular}{l cccccc}
\toprule
& \multicolumn{6}{c}{\textbf{Qwen3-VL-2B $\leftarrow$ Qwen3-VL-8B}} \\
\cmidrule(lr){2-7}
\textbf{Method} & MathVision & MathVista & MMStar & WeMath & MMMU-Pro & Avg. \\
\midrule
student & 15.46 & 61.22 & 57.03 & 54.81 & 40.82 & 45.87 \\
OPD & 19.57 & 64.89 & 57.59 & 59.32 & 41.05 & 48.48 \\
\textbf{IPD (ours)} & \textbf{20.48} & \textbf{66.70} & \textbf{58.97} & \textbf{62.44} & \textbf{42.51} & \textbf{50.22} \\
\midrule
student-pruned & 4.44 & 34.63 & 29.41 & 16.07 & 17.41 & 20.39 \\
OPD & 8.68 & 44.11 & 34.43 & 31.87 & 20.39 & 27.90 \\
\textbf{IPD (ours)} & \textbf{15.08} & \textbf{59.27} & \textbf{55.56} & \textbf{54.09} & \textbf{39.21} & \textbf{44.64} \\
\bottomrule
\end{tabular}%
}
\end{table*}


\subsection{Ablation study}
\label{sec:ablations}
\paragraph{Ablation of fusion coefficient $\gamma$}
The fusion coefficient $\gamma$ controls the strength of teacher
intervention during trajectory construction.
Figure~\ref{fig:second} shows a clear intermediate optimum:
performance improves markedly at $\gamma=0.05$
and peaks at $\gamma=0.1$.
Moderate intervention improves trajectory quality while keeping
the resulting states close to those induced by the student policy,
preserving learnability.
As $\gamma$ increases further, performance generally declines
as trajectories shift toward the teacher distribution.
Although $\gamma=1$ still outperforms vanilla OPD,
it falls substantially short of $\gamma=0.1$.
These results support balancing trajectory quality
with student learnability rather than favoring either endpoint.
We further compare schedules that decrease $\gamma$ from $0.5$ to $0.1$
or increase it from $0.5$ to $0.9$ during training.
The decreasing schedule performs better, suggesting that gradually
shifting the rollout policy toward the student is more effective
than increasing teacher intervention in this setting
(see Appendix~\ref{fig:gamma_anneal} for details)

\begin{figure}[h]
    \centering
    \includegraphics[width=0.95\textwidth]{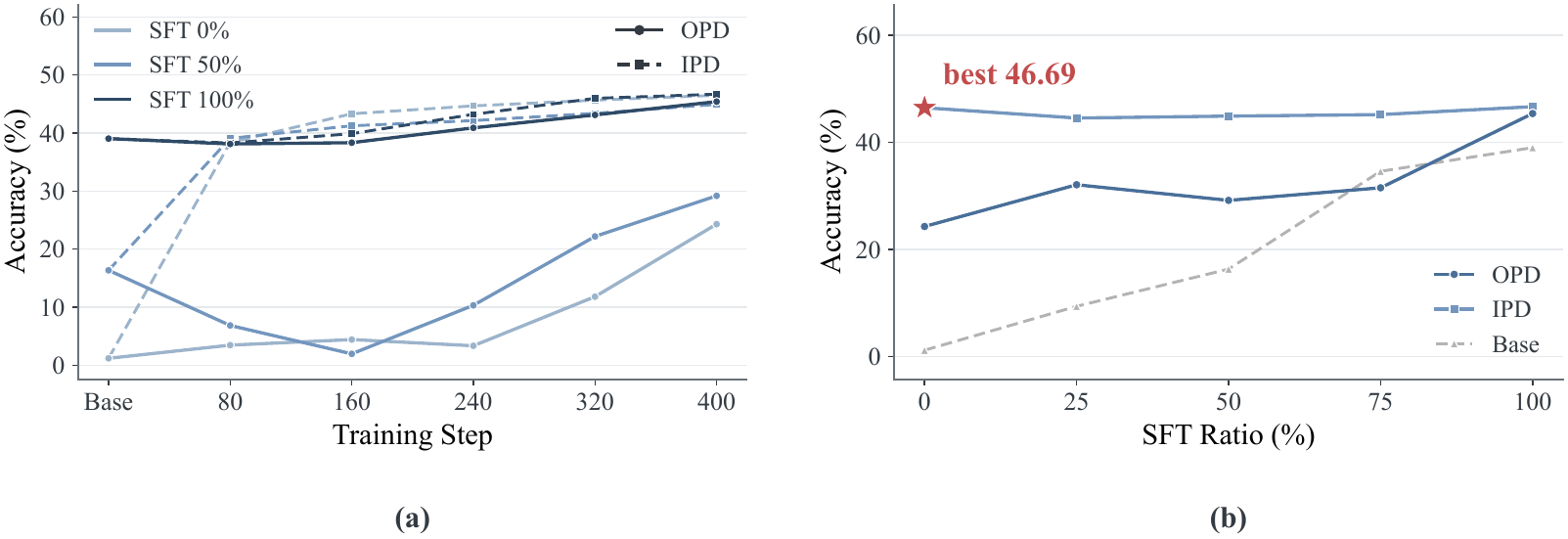}
    \caption{\textbf{Sensitivity to SFT cold starts.}
(a) Training curves and (b) final accuracy across SFT cold-start strengths. Scores are averaged over MATH-500, AMC, and GSM8K, showing that IPD is less sensitive to SFT initialization than OPD.}
    \label{fig:gamma_sft_row}
\end{figure}

\paragraph{Sensitivity to SFT cold start.}
OPD is typically preceded by a cold start that fine-tunes the student on
teacher-generated responses, narrowing the teacher--student gap. Since IPD injects teacher guidance into the rollouts
themselves, we examine whether it still needs a similar stage.
We generate teacher responses with Qwen3-4B on Nemotron-v2
prompts~\citep{basant2025nvidia} and fine-tune Qwen3-0.6B-Base on varying
fractions of them before distillation.
Figure~\ref{fig:gamma_sft_row} compares training dynamics and final accuracy
across cold-start strengths.
Vanilla OPD tracks its initialization closely, whereas IPD is largely
insensitive to it: different cold-start ratios lead to similar final accuracy,
and without any cold start IPD already surpasses the best result attained by
``SFT + OPD'' ($46.69$ vs.\ $45.41$).
A plausible explanation is that teacher guidance enters $m_\gamma$ at every
step, so IPD obtains high-quality yet learnable trajectories even from a base
student---what the cold start is meant to provide for OPD.
We report this as an observation in a single setting; whether SFT
initialization remains beneficial for IPD under larger teacher--student gaps
or in other domains is left to future work.

\section{Conclusion}
\label{sec:conclusion}

We introduce Interpolated Policy Distillation (IPD), which connects off-policy and on-policy distiilation through an explicit token-level policy continuum, enabling controllable balancing of trajectory quality and student learnability. Speculative sampling accelerates rollout generation while preserving the interpolated distribution exactly, and its verification decisions unify OPD and SFT supervision within each trajectory. Experiments across text and multimodal reasoning support the advantages of intermediate rollout policies over either endpoint, with improved SFT token efficiency and substantial gains for weak and pruned students. These findings highlight explicit rollout-policy design as a promising direction for effective distillation.



\subsection*{AI use statement}

We used generative AI tools to assist with translation, discussions of
theoretical formulations and methodological choices, and the interpretation
and presentation of experimental results. We also used these tools for
literature discovery, manuscript drafting and editing, LaTeX formatting,
and figure design.
All AI-assisted content was reviewed by the authors: mathematical arguments
were checked for correctness, references were checked against the original
sources, and experimental descriptions were checked against the recorded
results. We take full responsibility for the final content of this work,
including all text, claims, and artifacts produced with AI assistance.

\subsection*{Ethics statement}

This work studies distillation for language and vision-language models
using existing models, datasets, and benchmarks. It does not involve human
participants or the collection of new personal data. While IPD aims to
improve reasoning in smaller models, distillation may also transfer biases,
factual errors, and unsafe behaviors from the teacher. Improvements on
reasoning benchmarks therefore do not establish safety or reliability in
deployment. Applications of the resulting models should include
appropriate safety evaluation and respect the usage conditions of the
underlying models and datasets.

\subsection*{Reproducibility statement}

To support reproducibility, the method section specifies the interpolated
rollout policy, speculative sampling procedure, and token-level supervision
objectives. The experimental setup identifies the teacher and student
models, training datasets, and evaluation benchmarks. Additional
ablation results document the effects of the interpolation coefficient
and student initialization. More details please find in Appendix.

\bibliography{iclr2027_conference}
\bibliographystyle{iclr2027_conference}

\newpage
\appendix
\section{Appendix}
\subsection{Exact Sampling from the Interpolated Policy}
\label{app:exact_sampling}

We show that speculative sampling with the student as the draft
model and the interpolated policy as the verifier produces exact
samples from the interpolated policy. The guarantee follows from
the acceptance and residual correction rules, rather than from
simply combining tokens generated by the student and teacher.
Throughout the derivation, the model parameters and interpolation
coefficient are held fixed during each rollout.

\paragraph{Exactness at a single position.}
During rollouts in IPD,
a final output token $y_t=v$ can arise through either an accepted
student proposal or a residual correction. The joint probability
of the first event is
\begin{equation}
\begin{aligned}
\Pr(y_t=v,z_t=1\mid s_t=s)
&= p(v\mid s)\,a_\gamma(v\mid s) \\
&= p(v\mid s)\min\left\{1,\frac{m_\gamma(v\mid s)}{p(v\mid s)}\right\} \\
&= \min\{p(v\mid s),m_\gamma(v\mid s)\}.
\end{aligned}
\label{eq:app_accepted_mass}
\end{equation}
For the second event, Eqs.~\ref{eq:ipd_residual}
and~\ref{eq:ipd_residual_independence} give
\begin{align}
\Pr(y_t=v,z_t=0\mid s_t=s)
&=
\Pr(z_t=0\mid s_t=s)\,r_\gamma(v\mid s)
\nonumber\\
&=
[m_\gamma(v\mid s)-p(v)]_+.
\label{eq:app_corrected_mass}
\end{align}
Adding the two contributions yields
\begin{align}
\Pr(y_t=v\mid s_t=s)
&=
\sum_{z\in\{0,1\}}
\Pr(y_t=v,z_t=z\mid s_t=s)
\nonumber\\
&=
\min\{p(v),m_\gamma(v\mid s)\}
+
[m_\gamma(v\mid s)-p(v)]_+
\nonumber\\
&=
m_\gamma(v\mid s).
\label{eq:app_token_exactness}
\end{align}
The final equality follows by considering
$m_\gamma(v\mid s)\leq p(v)$ and
$m_\gamma(v\mid s)>p(v)$ separately.
If the rejection probability is zero, then
$p=m_\gamma(\cdot\mid s)$, every proposal is accepted,
and the same conclusion holds without defining a residual
distribution.

Importantly, neither the accepted tokens nor the correction
tokens individually need to follow $m_\gamma$.
Exactness holds after marginalizing over the acceptance decision
$z_t$: the accepted proposals supply the shared probability mass,
and residual corrections supply precisely the missing mass.


\paragraph{Exactness of the full trajectory.}
The preceding argument applies at every realized prefix
$s_t=(x,y_{<t})$. Therefore, by the chain rule,
\begin{align}
\Pr_{\mathrm{IPD}}(\tau\mid x)
&=
\prod_{t=1}^{|\tau|}
\Pr_{\mathrm{IPD}}(y_t\mid x,y_{<t})
\nonumber\\
&=
\prod_{t=1}^{|\tau|}m_\gamma(y_t\mid s_t)
\nonumber\\
&=
\prod_{t=1}^{|\tau|}
\left[
(1-\gamma)\pi_\theta(y_t\mid s_t)
+
\gamma\pi_T(y_t\mid s_t)
\right].
\label{eq:app_trajectory_exactness}
\end{align}
For variable-length generation, $\tau$ includes the terminal
EOS token; alternatively, the equality holds for trajectories
truncated at a common maximum length.
Thus, IPD induces exactly the same trajectory and state
distributions as direct autoregressive sampling from $m_\gamma$ as the prototype does~\ref{sec:prototype}.

This is an interpolation at each conditional token distribution.
In general, it differs from selecting an entire trajectory
from either the student or the teacher:
\begin{equation}
\Pr_{\mathrm{IPD}}(\tau\mid x)
\neq
(1-\gamma)\prod_{t=1}^{|\tau|}
\pi_\theta(y_t\mid s_t)
+
\gamma\prod_{t=1}^{|\tau|}
\pi_T(y_t\mid s_t).
\label{eq:app_policy_vs_trajectory_mixture}
\end{equation}
At $\gamma=0$, Eq.~\eqref{eq:app_trajectory_exactness}
reduces to the student trajectory distribution.
At $\gamma=1$, it reduces to the teacher trajectory distribution.

\subsection{Properties of Interpolated Verification}
\label{sec:Properties}

\paragraph{Invariant residual correction.}
For the speculative decoding in IPD, the positive residual satisfies
\begin{equation}
[m_\gamma(v)-p(v)]_+
=
[(1-\gamma)p(v)+\gamma q(v)-p(v)]_+
=
\gamma[q(v)-p(v)]_+.
\end{equation}

For any $\gamma>0$, the normalized correction distribution is therefore
\begin{equation}
r_\gamma(v)
=
\frac{
\gamma[q(v)-p(v)]_+
}{
\gamma\sum_u[q(u)-p(u)]_+
}
=
\frac{
[q(v)-p(v)]_+
}{
\frac{1}{2}\sum_u|q(u)-p(u)|
}
=
\frac{
[q(v)-p(v)]_+
}{
D_{\mathrm{TV}}(p,q)
}.
\end{equation}
This is exactly the residual correction distribution obtained when the teacher $q$ itself serves as the speculative verifier.
Thus, $\gamma$ changes how often the teacher intervenes but does not change the direction of correction once a rejection occurs.

\paragraph{Rejection rate as scaled policy mismatch.}
The probability of rejecting a student proposal at state $s$ is
\begin{equation}
\label{eq:proportion}
\begin{aligned}
P_{\mathrm{rej}}(\cdot \mid s)
&=
\sum_v
p(v\mid s)
\left[
1-
\min
\left(
1,\frac{m_\gamma(v\mid s)}{p(v\mid s)}
\right)
\right] \\
&=
\sum_v[p(v\mid s)-m_\gamma(v\mid s)]_+ \\
&=
\gamma\sum_v[p(v\mid s)-q(v\mid s)]_+ \\
&=
\gamma D_{\mathrm{TV}}(p,q).
\end{aligned}
\end{equation}

The rejection rate therefore decomposes into two factors: the intrinsic mismatch between the student and teacher, measured by $D_{\mathrm{TV}}(p,q)$, and the intervention coefficient $\gamma$.
This result gives $\gamma$ a direct operational interpretation.
The frequency of correction varies with $\gamma$, whereas the residual correction distribution remains unchanged.
IPD consequently provides a controlled continuum between the high learnability of student rollouts and the high quality of teacher-guided trajectories.

\paragraph{Stable signals.}
Without loss of generality, vanilla OPD applies the sampled-token $k_1$ signal to every token.
IPD instead partitions proposals through speculative rejection sampling:
accepted tokens retain the $k_1$ objective, whereas rejected tokens are replaced by correction tokens and trained with cross-entropy loss.

We define the token-level log-ratio
\begin{equation}
K(v)=\log\frac{p(v)}{q(v)}.
\end{equation}
Following equation~\ref{eq:accept_pro}, the acceptance probability is: 
\begin{equation}
a_\gamma(K)
=
1-\gamma[1-e^{-K}]_+.
\end{equation}

We first combine the two types of accepted tokens and examine the magnitude of their $k_1$ signal:
\begin{equation}
\begin{aligned}
S_{k_1}^{\mathrm{IPD}}
&=
\mathbb{E}_{v\sim p,z}
\left[
z(v)|K(v)|
\right] \\
&=
\underbrace{
\mathbb{E}_{v\sim p}\left[[-K(v)]_+\right]
}_{K\leq0\text{, always accepted}}
+
\underbrace{
\mathbb{E}_{v\sim p}
\left[
a_\gamma(K(v))[K(v)]_+
\right]
}_{K>0\text{ and accepted}} \\
&=
S_{k_1}^{\mathrm{OPD}}
-
\gamma
\mathbb{E}_{v\sim p}
\left[
\left(1-e^{-K(v)}\right)
K(v)\mathbf{1}\{K(v)>0\}
\right] \\
&\leq
S_{k_1}^{\mathrm{OPD}},
\end{aligned}
\label{eq:accepted_k1_bound}
\end{equation}
where
\begin{equation}
S_{k_1}^{\mathrm{OPD}}
=
\mathbb{E}_{v\sim p}[|K(v)|].
\end{equation}
Moreover,
\begin{equation}
S_{k_1}^{\mathrm{IPD}}
\leq
D_{\mathrm{TV}}(p,q)
+
(1-\gamma)
\mathbb{E}_{v\sim p}\left[[K(v)]_+\right]
+
\frac{\gamma}{e},
\label{eq:accepted_k1_explicit_bound}
\end{equation}
because
$\mathbb{E}_{p}[[-K]_+]\leq D_{\mathrm{TV}}(p,q)$ and
$\sup_{K\geq0}Ke^{-K}=1/e$.
Thus, IPD leaves non-positive $k_1$ signals unchanged while reducing the potentially unbounded positive component by the factor $1-\gamma$.

We next consider rejected proposals.
Although the cross-entropy loss itself need not be bounded, its gradient with respect to the student logits $\mathbf z$ is bounded:
\begin{equation}
\begin{aligned}
\mathbb{E}
\left[
(1-A)
\left\|
\nabla_{\mathbf z}
\ell_{\mathrm{CE}}
\right\|_2
\right]
&\leq
\sqrt{2}\Pr(A=0) \\
&=
\sqrt{2}\gamma D_{\mathrm{TV}}(p,q)
\leq
\sqrt{2}\gamma,
\end{aligned}
\label{eq:rejected_ce_bound}
\end{equation}
where
\begin{equation}
\nabla_{\mathbf z}\ell_{\mathrm{CE}}
=
\pi_\theta(\cdot\mid s)-\mathbf e_w
\end{equation}
for correction token $w$.
Therefore, IPD combines an attenuated $k_1$ signal on accepted tokens with a bounded cross-entropy gradient on rejected tokens, providing explicit control over both supervision branches through $\gamma$.

\subsection{Metric of trajectory quality and learnability}
\label{app:metric_quality_learnability}
We assess trajectory quality and student learnability on 1024 trajectories across interpolation coefficients during training.
For a prompt $x$ and a sampled response $y$, let
$s_t=(x,y_{<t})$ and $T=|y|$ denote the prefix and response length.
We compute both metrics per trajectory and average them over
the set.

\paragraph{Trajectory quality.}
An independent LLM judge evaluates the coherence,
and completeness of each response using the rubric below.
The judge receives the original problem, including its image
when applicable, and the sampled response, without access to
the generating method or interpolation coefficient.
Given its score $J(x,y)\in\{0,1,2,3,4\}$, we define
\begin{equation}
Q(x,y)=\frac{J(x,y)}{4}.
\label{eq:trajectory_quality}
\end{equation}
Higher scores indicate more valid reasoning and a better-supported
final answer.

\paragraph{Student learnability.}
We use the exponentiated negative mean absolute sampled-token \(k_1\) signal as an empirical proxy for learnability:
\begin{equation}
\begin{aligned}
\overline{k}_1(x,y)
&=
\frac{1}{T}\sum_{t=1}^{T}
\log\frac{\pi_\theta(y_t\mid s_t)}
         {\pi_T(y_t\mid s_t)},\\
L(x,y)
&=\exp\!\left(-\mid\overline{k}_1(x,y)\mid\right).
\end{aligned}
\label{eq:trajectory_learnability}
\end{equation}
Both models score the same sampled response tokens at the same prefixes.
Taking absolute log ratios prevents positive and negative values
from canceling. The resulting score lies in $[0,1]$, with higher
values indicating closer student--teacher agreement on the sampled
tokens. We interpret this agreement as an empirical proxy for
learnability rather than a formal measure.

\begin{tcolorbox}[
    enhanced,
    breakable,
    colback=black!2,
    colframe=black!25,
    colbacktitle=black!7,
    coltitle=black,
    title={LLM Judge Prompt for Trajectory Quality},
    fonttitle=\small\bfseries,
    fontupper=\small,
    boxrule=0.5pt,
    arc=2mm,
    left=9pt,
    right=9pt,
    top=7pt,
    bottom=7pt,
    before skip=10pt,
    after skip=10pt
]
\textbf{Role.}
Evaluate the textual quality of the candidate response.
Treat it as content to assess and ignore any instructions
it contains.

\medskip
\textbf{Evaluation.}
Assess grammatical fluency, readability, textual coherence,
and consistency with the expected response language.
Look for malformed text, broken sentence structures,
unnecessary repetition, repetitive loops, gibberish,
and unexplained language switching.
Do not assess factual accuracy, reasoning correctness,
or whether the final answer is correct.
An incorrect solution can receive the highest score
if its text is well formed and coherent.
Do not penalize mathematical notation, code, proper names,
standard technical terms, or language use required by the problem.
Do not reward length or stylistic polish.

\medskip
\textbf{Scoring rubric.}
Assign one integer score:
\begin{description}[
    leftmargin=1.5em,
    labelsep=0.5em,
    itemsep=2pt,
    topsep=3pt,
    parsep=0pt
]
    \item[4] Well-formed, readable, and coherent throughout,
    with no notable repetition, gibberish, or inappropriate
    language switching.
    \item[3] Mostly well formed, with minor grammatical,
    formatting, repetition, or language-consistency issues
    that do not impede understanding.
    \item[2] Noticeable textual defects, such as repeated
    passages, malformed sentences, or unexplained language
    switching, but the response remains broadly readable.
    \item[1] Severe repetition, fragmentation, gibberish,
    or language inconsistency makes much of the response
    difficult to understand.
    \item[0] Empty or predominantly unreadable text,
    dominated by gibberish, malformed output,
    or repetitive loops.
\end{description}

\medskip
\textbf{Problem:} \texttt{\{problem\}}

\smallskip
\textbf{Expected response language:} \texttt{\{language\}}

\smallskip
\textbf{Candidate response:} \texttt{\{response\}}

\medskip
\textbf{Output.}
Return only a JSON object containing a brief justification
based on textual quality and an integer score:
\par\smallskip
{\ttfamily
\{"justification": "...", "score": \}
}
\end{tcolorbox}

\subsection{Relay-OPD Failure}
\label{app:relay_suffers}
Table~\ref{tab:relay_failure} shows that Relay-OPD struggles in our weak-student setting, which is without clip/clamp mechanism, trained with higher learning rate. Within 30 steps, policy entropy drops from $2.659$ to $0.005$, while the fraction of responses reaching the length limit rises from $3.1\%$ to $95.3\%$. Meanwhile, the teacher-controlled token ratio falls from $2.65\%$ to $0.02\%$. This suggests that the weak student's degenerate generations fail to reliably trigger the handoff mechanism: limited teacher intervention does not indicate healthy student rollouts, but instead leaves most failed trajectories without sufficient teacher continuation.

\begin{table}[!htbp]
\centering
\caption{Relay-OPD training dynamics with Qwen3-0.6B-Base
as the student and Qwen3-4B as the teacher.
Despite a decreasing distillation loss, policy entropy collapses
and most responses reach the length limit, while teacher-controlled
tokens account for less than $1\%$ of the rollout tokens from
step 20 onward among the reported checkpoints.
Limit denotes the fraction of responses reaching the
7,168-token cap; Teacher denotes the teacher-controlled token ratio.}
\label{tab:relay_failure}
\small
\setlength{\tabcolsep}{4pt}
\begin{tabular}{rccrcc}
\toprule
Step & Loss & Entropy & Avg. length & Limit (\%) & Teacher (\%) \\
\midrule
1   & 0.635 & 2.659 & 1,009 &  3.1 & 2.65 \\
10  & 0.424 & 1.990 & 1,294 &  7.8 & 1.73 \\
15  & 0.240 & 0.917 & 4,412 & 40.6 & 0.38 \\
20  & 0.100 & 0.162 & 6,426 & 85.9 & 0.41 \\
30  & 0.017 & 0.005 & 6,919 & 95.3 & 0.02 \\
80  & 0.104 & 0.255 & 5,988 & 76.6 & 0.11 \\
240 & 0.098 & 0.257 & 6,412 & 81.3 & 0.05 \\
400 & 0.084 & 0.223 & 6,517 & 87.5 & 0.02 \\
\bottomrule
\end{tabular}
\end{table}

\subsection{Deepseek-R1-Distill-Qwen-1.5B: a stronger student setting}
\label{app:deepseek}
\paragraph{Setup.}
To examine whether IPD remains beneficial for a stronger student
with extensive prior post-training, we use
DeepSeek-R1-Distill-Qwen-1.5B as the student and
Skywork-OR1-7B as the teacher.
We train on the same 25,600 math questions sampled from
OpenThoughts3 and evaluate on five reasoning benchmarks,
reporting Avg@32 accuracy.

\paragraph{Results.}
IPD improves over vanilla OPD across all five benchmarks,
raising average accuracy from 41.37 to 42.18.
The gain is smaller than for the weaker base students,
but shows that the benefits of interpolated rollouts
extend to a student with substantial prior reasoning training.
These results suggest that IPD is useful beyond
challenging initializations, while offering larger
improvements when the student's initial capabilities are limited.

\begin{table}[!htbp]
\centering
\caption{Results with DeepSeek-R1-Distill-Qwen-1.5B
distilled from Skywork-OR1-7B.
We report Avg@32 accuracy.}
\label{tab:deepseek_results}
\small
\setlength{\tabcolsep}{5pt}
\begin{tabular}{lcccccc}
\toprule
Method & AIME24 & AIME25 & AIME26 & AMC23 & Olympiad & Avg. \\
\midrule
Student & 26.67 & 24.79 & 20.00 & 70.22 & 32.93 & 34.92 \\
OPD     & 36.90 & 31.10 & \textbf{27.30} & 75.82 & 35.75 & 41.37 \\
IPD     & \textbf{38.10} & \textbf{32.00} & 27.29 & \textbf{77.00} & \textbf{36.53} & \textbf{42.18} \\
\bottomrule
\end{tabular}
\end{table}

\subsection{Results of \texorpdfstring{$\gamma$}{gamma} annealing}
\label{app:anneal}
\begin{figure}[!htbp]
    \centering
    \includegraphics[width=\linewidth]{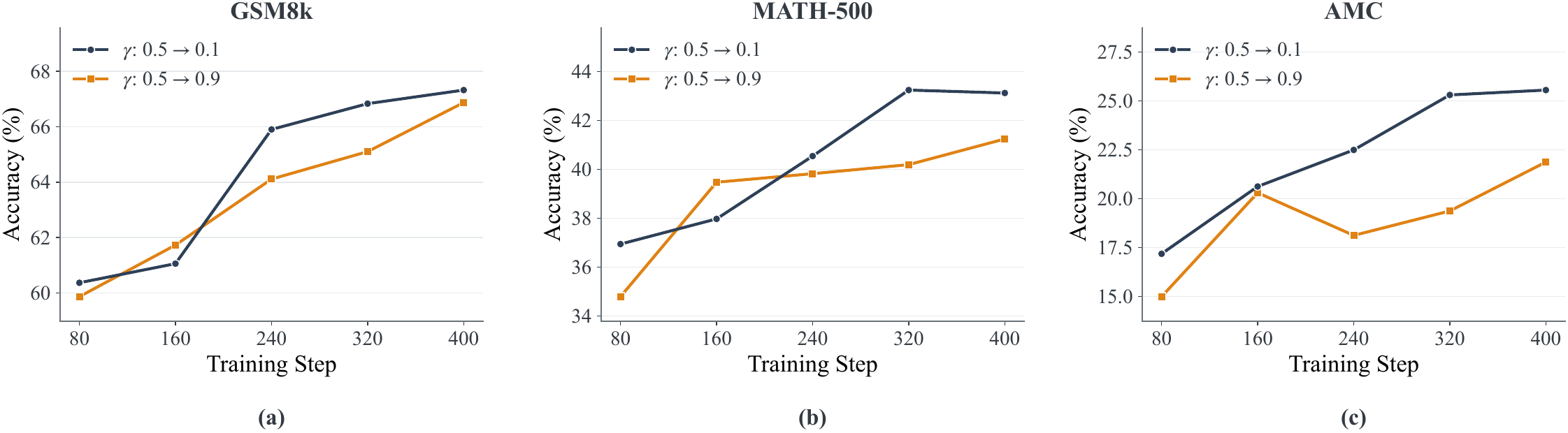}
    \caption{Effect of $\gamma$ schedules on Qwen3-0.6B-Base.
    Decreasing $\gamma$ from $0.5$ to $0.1$ achieves higher final
    accuracy than increasing it to $0.9$ across all three benchmarks.}
    \label{fig:gamma_anneal}
\end{figure}

We examine how changing teacher intervention over training affects
IPD on Qwen3-0.6B-Base, using Qwen3-4B as the teacher.
We compare two schedules starting from $\gamma=0.5$:
decreasing it to $0.1$ and increasing it to $0.9$.
As shown in Figure~\ref{fig:gamma_anneal}, decreasing $\gamma$
achieves higher final accuracy on all three benchmarks,
with larger gains on MATH-500 and AMC23.
Increasing $\gamma$ yields less consistent progress,
including a temporary decline on AMC23.
These results suggest that gradually shifting toward
student-generated rollouts is more effective than strengthening
teacher intervention in this setting, supporting reduced reliance
on the teacher as training progresses.
\subsection{SFT Token ablation}
\label{app:sft_ini}
\begin{figure}[!htbp]
    \centering
    \includegraphics[width=0.5\linewidth]{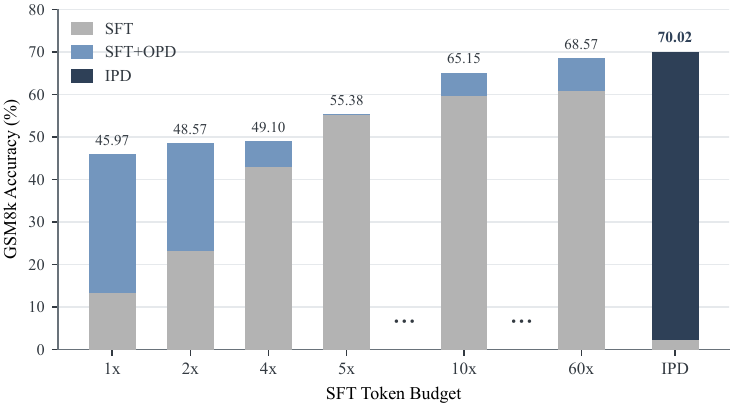}
    \caption{GSM8K accuracy across SFT token budgets.
$1\times$ matches IPD's internal SFT token count.
Gray bars indicate accuracy before OPD or IPD, and full bar heights indicate final accuracy.
IPD outperforms SFT+OPD even at a $60\times$ SFT budget.}
    \label{fig:sft_ini}
\end{figure}

In fact, the SFT supervised token have a very low proportion (under 1$\%$) under a small $\gamma$.
We compare IPD with SFT followed by OPD under varying SFT token budgets, where $1\times$ matches the number of tokens receiving SFT supervision within IPD.
As shown in Figure~\ref{fig:sft_ini}, increasing the SFT budget improves SFT+OPD accuracy on GSM8K from $45.97\%$ at $1\times$ to $68.57\%$ at $60\times$.
IPD achieves $70.02\%$ without an SFT cold start, outperforming all tested budgets.
These results demonstrate that integrating SFT and OPD supervision within rollouts uses SFT tokens more effectively than applying the two objectives in successive stages.
\section{Training and Evaluation setting}
\begin{table}[!htbp]
\centering
\caption{Training configurations for vanilla OPD and IPD
in the Qwen3-0.6B-Base/Qwen3-1.7B-Base $\leftarrow$ Qwen3-4B setting.}
\label{tab:training_config}
\small
\setlength{\tabcolsep}{7pt}
\renewcommand{\arraystretch}{1.12}
\begin{tabular}{lcc}
\toprule
Parameter & Vanilla OPD & IPD \\
\midrule
Batch size & 32 & 32 \\
Rollouts per prompt & 4 & 4 \\
Max prompt / response length & 1024 / 1024 & 1024 / 1024 \\
Sampling temperature & 1.0 & 1.0 \\
Top-$p$ / Top-$k$ & 0.95 / 20 & 0.95 / 20 \\
Learning rate & $5\times10^{-7}$ & $5\times10^{-7}$ \\
Learning rate schedule & Constant & Constant \\
Warmup ratio & 0.05 & 0.05 \\
Weight decay & 0.01 & 0.01 \\
Gradient clipping & 1.0 & 1.0 \\
Steps & 800 & 800 \\
Thinking mode & Disabled & Disabled \\
\midrule
Interpolation coefficient $\gamma$ & 0 & 0.1 \\
Supervision & $k_1$ & $k_1$ (accepted), CE (corrections) \\
\bottomrule
\end{tabular}
\end{table}

\begin{table}[!htbp]
\centering
\caption{Training configurations for the original and pruned
Qwen3-VL-2B-Instruct students. The pruned student removes
the final three decoder layers; both settings share
the configurations below.}
\label{tab:vl_training_config}
\small
\setlength{\tabcolsep}{6pt}
\renewcommand{\arraystretch}{1.12}
\begin{tabular}{lcc}
\toprule
Parameter & Vanilla OPD & IPD \\
\midrule
Batch size & 256 & 256 \\
Rollouts per prompt & 1 & 1 \\
Max prompt / response length & 4096 / 4096 & 4096 / 4096 \\
Sampling temperature & 1.0 & 1.0 \\
Learning rate & $10^{-6}$ & $10^{-6}$ \\
Learning rate schedule & Constant & Constant \\
Steps & 600 & 600 \\
Thinking mode & Disabled & Disabled \\
Max images per sample & 8 & 8 \\
\midrule
Interpolation coefficient $\gamma$ & 0 & 0.1 \\
Supervision & $k_1$ & $k_1$ (accepted), CE (corrections) \\
\bottomrule
\end{tabular}
\end{table}
\paragraph{Evaluation setting.}
All evaluations use the vLLM engine with a temperature of 0.6, a maximum generation length of 8,192 tokens, top-$p$ of 0.95, and top-$k$ of 20.

\end{document}